\documentclass[A4paper, 11 pt, conference]{ieeeconf}  
\IEEEoverridecommandlockouts       
\usepackage{geometry}
\usepackage{url}
\usepackage{graphics} 
\usepackage{epsfig} 
\usepackage{mathptmx} 
\usepackage{times} 
\usepackage{amsmath} 
\usepackage{amssymb}  
\usepackage{subfigure}
\usepackage{algorithm} 
\usepackage{algorithmic} 
\usepackage{multirow} 
\begin{document}

\title{\LARGE \bf Electric Vehicle Automatic Charging System Based on Vision-force Fusion}

\author{Dashun Guo, Liang Xie, Hongxiang Yu, Yue Wang and Rong Xiong 
\thanks{
   Yue Wang is the corresponding author and with Control Science of Engineering, Zhejiang University, Hangzhou, Zhejiang, China, {\tt\small ywang24zju.edu.cn}.
}}

\maketitle 
\thispagestyle{empty}

\begin{abstract}
Electric vehicles are an emerging means of transportation with environmental friendliness. The automatic charging is a hot topic in this field that is full of challenges. We introduce a complete automatic charging system based on vision-force fusion, which includes perception, planning and control for robot manipulations of the system. We design the whole system in simulation and transfer it to the real world. The experimental results prove the effectiveness of our system.
\end{abstract}

\section{Introduction}
Electric vehicles are gaining more and more popularity due to their stability, high energy efficiency and environmental friendliness. The demand of achieving automatic charging of electric vehicles is rising in modern days. 

The challenges of the automatic charging can be summarized as (1) estimating the pose of the charging cover and opening it (2) estimating the pose of the charging port and plugging the charger into the port. In this paper, we introduce a complete system that operates in stages. In the first stage, the charging cover pose is estimated based on hybrid vision-force fusion from multi-sensors. Then the motion of the robot end effector is planned to open the cover to pave the way for the following manipulations. It is import to note that the proposed method is easy to complete without collecting real world datasets or consuming large amount of time for training compared with the recent learning-based methods \cite{6739623, he20173d}. In the second stage, the challenge can be generalized as a 6-DoF peg-in-hole task. We use the hybrid vision-force modality for the task as well for the reason that the vision-based peg-in-hole is fast with large error but imprecision due to the camera resolution, while the force-based peg-in-hole can achieve high precision but the initial error is limited within small range. In addition, vision can help achieve 6-DoF align while force-based method can only recognize the position error but hardly the orientation error, which is only 3-DoF. On this account, we propose a hybrid vision-force based approach to achieve the 6-DoF peg-in-hole with high precision, accuracy and efficiency. The whole system is trained totally in simulation by domain randomization which can be directly transferred to the open world tasks without any fine-tuning. A set of experiments are designed to verify the feasibility and effectiveness of the whole system both in simulation and real world environment. 

Our contributions are as follows:
\begin{enumerate}
    \item System: Propose a complete method including perception, planning, and control for the electric vehicle automatic charging system.
    \item Sensor Fusion: Propose a hybrid vision-force modality for complex manipulation tasks. 
    \item Sim2real: The whole system is trained in simulation and directly transferred to the real world without any fine-tuning.
\end{enumerate}


\section{Related Work}

\subsection{3D Pose Estimation}
Object pose estimation is the key to implement object manipulation. Drost and et al. present a 3D object recognition method called PPF \cite{drost2010model}. They build a hash table offline to store all four-dimensional features of the model as an overall description of the model and then match the parameter space online by Hoff voting to obtain a reliable pose. Choi and et al. improved the PPF by expanding the traditional four-dimensional geometric point pair features with color information to obtain CPPF, which is a ten-dimensional feature descriptor with RGB information added to each of the two points \cite{choi20123d}. They got better results than the original PPF are obtained on the color dataset. Tuzel and et al. proposed a maximal edge learning framework to recognize recognition features on the surface of 3D objects \cite{tuzel2014learning}. The algorithm selects and ranks features based on their importance in a given task, thereby improving accuracy and reducing computational costs. However, the above methods not only require the 3d model of the object is  as known, but also require a large labeled dataset in the real world and consuming time for training, which is considerably costly.
\begin{figure}[t]
\centering
\includegraphics[scale=0.55]{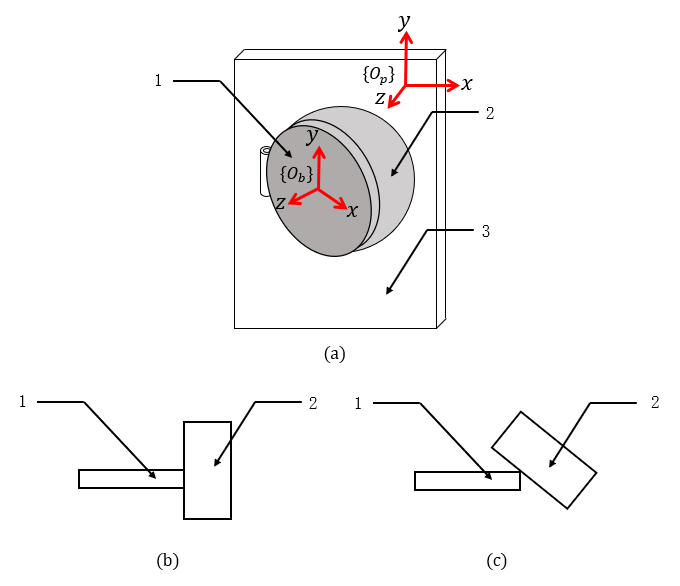}
\caption{Preparation for charging cover perception and manipulation. Fig.(a) shows the location relationship among the charging cover, charging port and the plane (denoted by $1, 2, 3$ respectively). The $\{O_p\}$ and $\{O_b\}$ represent the plane and the cover coordinate system. Fig.(b) and (c) show the ${H_{be}}_1$ and ${H_{be}}_2$ we use in the $attempt$ and $open$ stage. The $1,2$ represent the cover and the charger respectively.}
\label{pri_pos}
\end{figure}
\subsection{Peg-in-hole}
Contact-rich tasks, such as peg insertion, fastening screws have been studied for decades due to their relevance in manufacturing. Tradition methods for peg-in-hole task like spiral search and random search \cite{2020Fast} have been widely used for their simple realization and  generalization. However, the traditional methods can only solve peg-in-hole tasks in limited scenarios, such as small initial position error and no orientation error. Moreover, the accuracy and efficiency of these methods are not desirable in real world manufacturing. Recently, with the development of deep learning, the vision-based learning methods have achieve great progress in the robot manipulating fields \cite{2019Making}\cite{2019Form2Fit}\cite{2020Learning}\cite{2020Quickly}\cite{2021Tactile}\cite{2021Vision}. However, currently they have only been applied in simulation or real world with limited conditions.

\section{Method}
The whole system is calibrated. As shown in Fig. \ref{real_scene}, the F/T sensor is fixed at the robot wrist. The charger is fixed at the robot end effector. And the RGB-D camera with known intrinsic and extrinsic matrix is fixed in the hand. The charging port is fixed randomly at the world coordinate with a cover attached, and assume the surface of the port has rich features such as small circles for detection. The parameters of the cover are regarded as known, such as the cover's radius $r_b$, opening direction and the port's radius $r_o$.  Finally the task can be generalized as how to control the robot arm to plug the charger into the charging port by hybrid vision and force data.


\subsection{Charging Cover Perception and Manipulation}
\subsubsection{Preparation}
There are some indispensable preparations before the perception and manipulation. Assume that the coordinate system of the robot arm's base is the $world$ coordinate system $\{O_w\}$, firstly we need to get the rotation matrix of the plane $R_{wp}$ in $\{O_w\}$. The other preparation is to get the charger's poses in the cover coordinate system $\{O_b\}$, which are respectively ${H_{be}}_{1}$ in $attempt$ stage and ${H_{be}}_{2}$ in $open$ stage, shown in Fig.~\ref{pri_pos}. These preparations mentioned above can be implemented easily with the help of AprilTag.

\subsubsection{Perception}
We can obtain a point cloud $P_w$ via the RGB-D camera, which contains the point cloud $P_b$ representing the charging cover. A point cloud is represented as a set of 3D points $\{p_i | i=1 ,...,n\}$, where each point $p_i$ is a vector of its $(x ,y ,z)$ coordinate system plus extra feature channels such as color, normal etc \cite{qi2017pointnet}. The charging cover can be regard as a plane, that is, the points of the point cloud $P_b$ have the same normal. We cluster the same normals of $P_w$ by K-means so as to segment the $P_w$. And then 
by comparing the area of the plane and the cover, we can separate the $P_b$ from $P_w$.

The point is that due to the performance of the camera and the lighting conditions in the scene, some parts at the edge of the cover are missing in the $P_b$. It results in an accurate normal of the cover but a center point with error. We introduce an $attempt$ stage to correct this position error. As shown in Fig.~\ref{pri_pos}, specify that the direction of the cover's normal is the z-axis direction of the cover $\hat{z}_{cb}$ in the camera coordinate system $\{O_c\}$, and y-axis direction $\hat{y}_{cb}$ is the same as the plane $\hat{y}_{cp}$ in $\{O_c\}$, we can compute the rotation matrix of the cover $R_{cb}$ as follows:
\begin{equation}
\hat{x}_{cb} = \hat{y}_{cb}  \times \hat{z}_{cb} 
\label{rcb1}
\end{equation}
\begin{equation}
\hat{y}_{cb} = \hat{y}_{cp}
\label{rcb2}
\end{equation}
\begin{equation}
\hat{y}_{cp} = R_{cw} \hat{y}_{wp}
\label{rcb3}
\end{equation}
\begin{equation}
R_{cb}=\begin{bmatrix}
 \hat{x}_{cb}  &  \hat{y}_{cb} &  \hat{z}_{cb}
\end{bmatrix}
\label{rcb4}
\end{equation}
where $R_{cw}$ is the rotation matrix of the $world$ in $\{O_c\}$, which can obtained by calibration.
The next step is to control the charger to the ${H_{be}}_{1}$ we prepared. Due to the position error mentioned above, we set the charger to ${^bH{}'_{e}}_{1}$, which translates a distance $x_1$ with respect to ${H_{be}}_{1}$ in the x-axis direction. Then we slowly control the charger along the negative direction of the x-axis until the charger and cover contact, which we can inferred from the force data. Record the displacement $x_2$, then the error $x_e$ can be calculated as
\begin{equation}
x_e = -(x_2 - x_1)
\label{rcb5}
\end{equation}

At this point, we complete the pose estimation of the cover.
\begin{figure}[t]
\centerline{\includegraphics[width=3.5in]{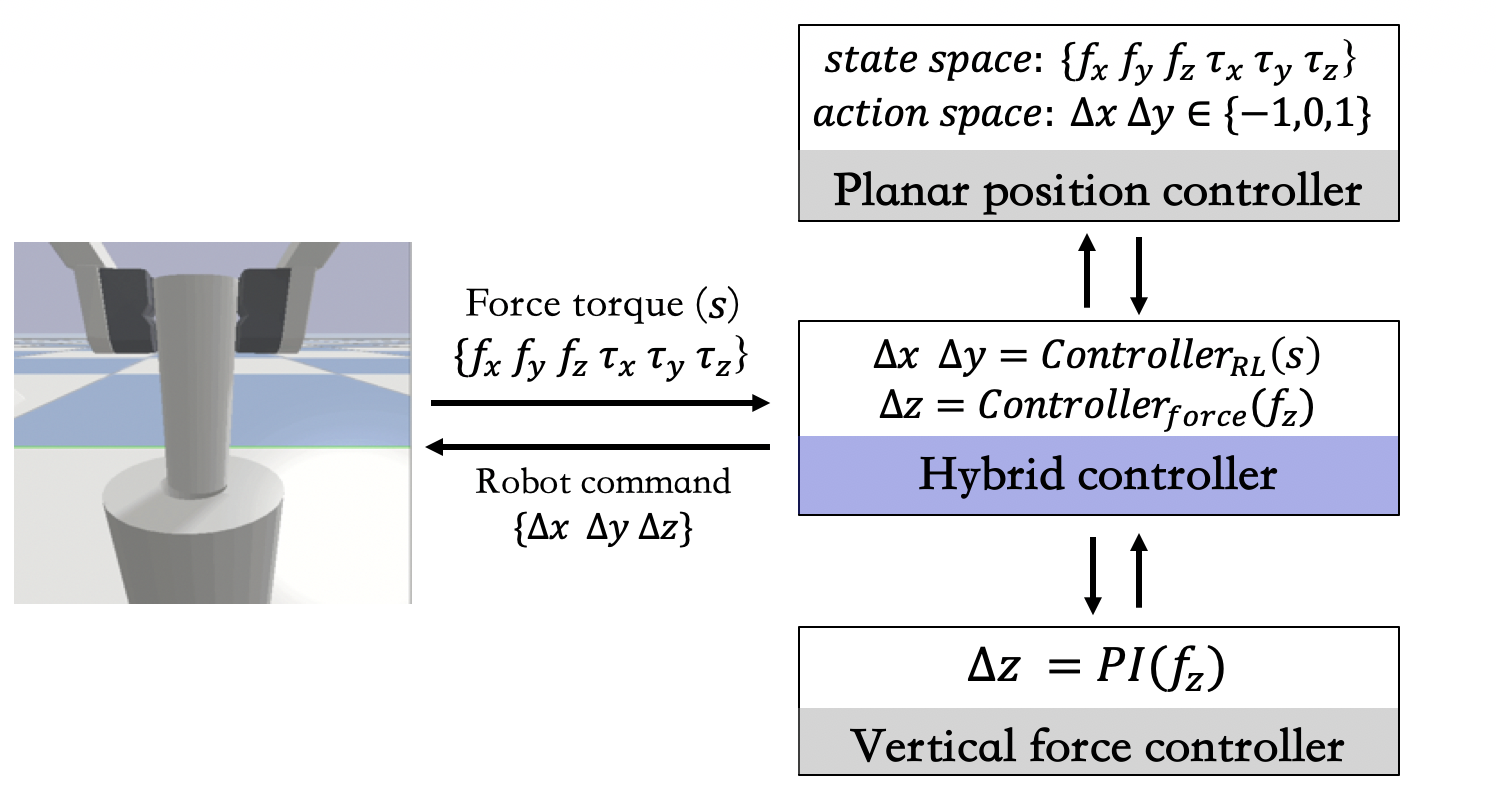}}
\caption{Hybrid position-force controller.}
\label{figg}
\end{figure}
\subsubsection{Manipulation}
The cover is hinged at the charging port and can be rotated around the hinge axis, that means, the charger should keep in contact with the cover and rotating around the hinge axis. The first step is control the charger to the ${H_{be}}_2$ to touch the cover. Secondly we need the pose of the axis $H_{ca}$ in $\{O_c\}$ to calculate the velocity of the charger. The rotation matrix of the axis $R_{ca}$ in $\{O_c\}$ is the same as the plane, and the center can be obtained from the center of the cover along the negative direction of its x-axis, so that we obtain $H_{ca}$. Assume the angular velocity of the cover $\omega_{d}$ is
\begin{equation}
\nonumber
\omega_{d} = \begin{bmatrix}
0   &  0 &  -0.1
\end{bmatrix}
\label{vel0}
\end{equation}
so the liner velocity of the cover $v_{d}$ is
\begin{equation}
\nonumber
v_{d} = \omega_d \times T_{we}
\label{vel0}
\end{equation}
where $T_{we}$ is the position of the end effector.
Then we control the charger by
\begin{equation}
v_e = R_{wc}R_{ca}v_{b} 
\label{vel1}
\end{equation}
\begin{equation}
\omega_e = R_{wc}R_{ca}v_{b} 
\label{vel2}
\end{equation}
\begin{equation}
R_{wc} = R_{we}R_{ec}
\label{vel3}
\end{equation}
where $R_{we}$ is the orientation of the end effector in $\{O_w\}$.
Force data from the wrist F/T sensor can help us to determine whether to stop the charger.

\subsection{Peg-in-Hole}
In this section, we explain the multi-modal two-stage method for the peg-in-hole task, where the charger is regarded as the peg and the charging port is the hole. In the first stage, the visual servo is proposed to quickly search the charging port under large 6-DoF error and roughly align the pose of the charger and the charging port. In the second stage, we use the hybrid position-force controller to compensate the remaining error with high precision control for the insert stage.

\subsubsection{Search}
Image-based visual servo compares the image signal detected in real time directly with the image signal of the target pose, and uses the obtained image error for closed-loop feedback control. Image-based visual servo requires extracting feature points on the image and constructing a relationship between the pixel coordinate system error of the feature points and the end velocity of the robot arm.

The normal charging port has rich features for us to extract. In our method, we choose HSV to extract the feature point.

Assume that feature point extracted from the image is $q$, and the velocity is $\dot{q}$. Specify $\{\hat{O}_c\}$ is the current camera coordinate system and $\{O_c\}$ for the previous moment, then according to the rotation translation relationship we have
\begin{equation}
T_{\hat{c}q} = T_{\hat{c}c} + R_{\hat{c}c}T_{cq}
\label{visio1}
\end{equation}
Derivative for both sides of (\ref{visio1}), then
\begin{equation}
v_{\hat{c}q} = v_{\hat{c}c} + \omega_{\hat{c}c}T_{cq} + R_{\hat{c}c}v_{cq}
\label{visio1}
\end{equation}
Then 
with the camera internal equation
\begin{equation}
u=f_x\frac{x}{z} 
\label{visio2}
\end{equation}
\begin{equation}
v=f_x\frac{y}{z} 
\label{visio2}
\end{equation}
where the $u, v$ are the coordinates of $p$ in the $pixel$ coordinate system $\{O_i\}$, $f_x, f_y$ are the focus of the camera and $x,y,z$ are the coordinates of $p$ in the camera coordinate system $\{O_c\}$, we can get the relationship between the velocity $v_i$ in $\{O_i\}$ as follows:
\begin{equation}
v_{ip} = J_pv_{\hat{c}c}
\label{visio3}
\end{equation}
\begin{equation}
J_p = \begin{bmatrix}
 \frac{f_x}{z}  & 0 & -\frac{u}{z}  & -\frac{uv}{f_y}  & \frac{f_x^2+u^2}{f_x}   & -\frac{f_xv}{f_y} \\
 0 & \frac{f_y}{z}  & -\frac{v}{z}   & -\frac{f_y^2+v^2}{f_y} & \frac{uv}{f_x} & \frac{f_yu}{f_x}
\end{bmatrix}
\label{visio4}
\end{equation}
where $v_{ip}$ is the velocity of $p$ in $\{O_i\}$, then we get the $v_{\hat{c}c}$ as
\begin{equation}
v_{\hat{c}c}=J_p^+v_{ip}
\label{visio5}
\end{equation}
Finally, convert $v_{\hat{c}c}$ to the velocity of the end effector $v_{we}$ in $\{O_w\}$ with
\begin{equation}
v_{we}=R_{we}R_{ec}v_{\hat{c}c}
\label{visio5}
\end{equation}
then we can control the charger with $v_{we}$ to align with the charging port.
\begin{figure}[t]
\centering
\includegraphics[scale=0.36]{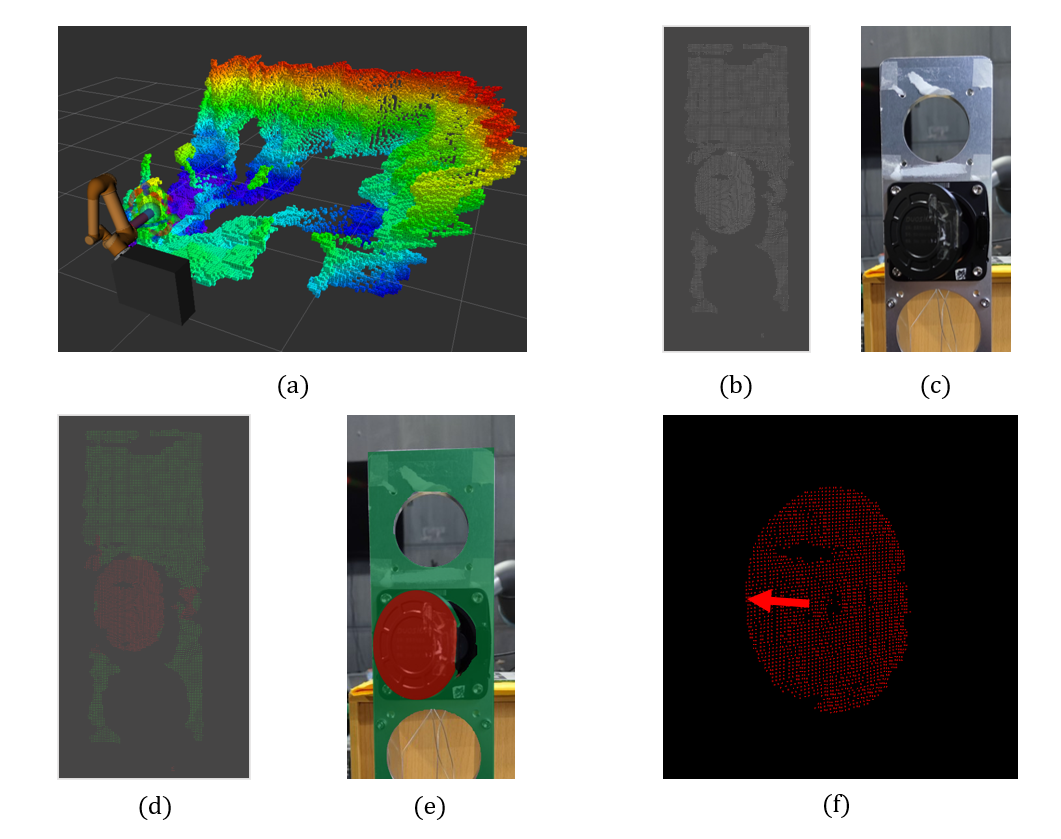}
\caption{Process of the charging cover perception. Fig.(a) is the world point cloud obtained from the RGB-D camera. Fig.(b) shows the cropped point cloud contained the plane and charging port shown in Fig.(c). Fig.(d) and (e) show the segmentation results by K-means. The part with green mask represent the plane and the red represent the cover. Fig.(f) is the final result, showing the center and the normal of the cover. }
\label{esti}
\end{figure}
\subsubsection{Insert}
When the charging port is searched and roughly aligned by visual servo, the hybrid position-force controller is used for the insert stage with the remaining error. As shown in Fig.~\ref{figg}, the insert task is generalized as commanding the robot end effector(peg) to the desired position(hole) with 3 degree-of-freedom that the planar position is learned by the RL algorithm, and the vertical position is calculated by the PI controller.

The RL controller observes the current state $s$ of the system defined as:

\begin{equation}
s = \{f_x, f_y, f_z, \tau_x, \tau_y, \tau_z\},
\label{eq}
\end{equation}
where $f$ and $\tau$ are the average force and torque obtained from the wrist F/T sensor; the subscript $x,y,z$ denotes the axis. Then the controller outputs an action $a$ commanding the end effector to move 1mm distance along anyone of the four axis direction in the planar space. We use A2C \cite{mnih2016asynchronous} algorithm as the RL controller which is enough for the current task. During learning, the initial relative position between the peg and the hole is randomized within 5mm, as the former visual servo stage can guarantee the position align with around 2mm error. At each time step, the RL controller receives the observation $s$ from the system and generates the action $a$ with random exploration, then a reward $r$ is obtained from the system. By interacting with the system over time, the RL controller strives to maximize the cumulative reward:

\begin{equation}
R_T = \sum_{t=0}^T \gamma^tr_t,
\label{eq}
\end{equation}
where $\gamma$ is the discount factor, and $T$ is the episode length. We use the sparse reward function as defined in (\ref{eqq})  that a reward is returned from the system only when the trial succeeds. The reward is designed to encourage the RL controller to finish the episode with minimal steps.

\begin{equation}
r = 1 - \dfrac{T}{T_{max}}
\label{eqq}
\end{equation}

We use PI controller for force control to get the vertical displacement command as defined in (\ref{eqqq}). 
\begin{equation}
dz = PI(f_z)
\label{eqqq}
\end{equation}
The force in z-axis is maintained as 10N while the charger makes contact with the charging port and moves in the planar plane. The insert is regarded as successful when $f_z$ is lower than the preset thresholds. 

Finally, the closed loop control with F/T feedback is achieved by the hybrid position-force controller.

\begin{figure}[t]
\centering
\includegraphics[scale=0.4]{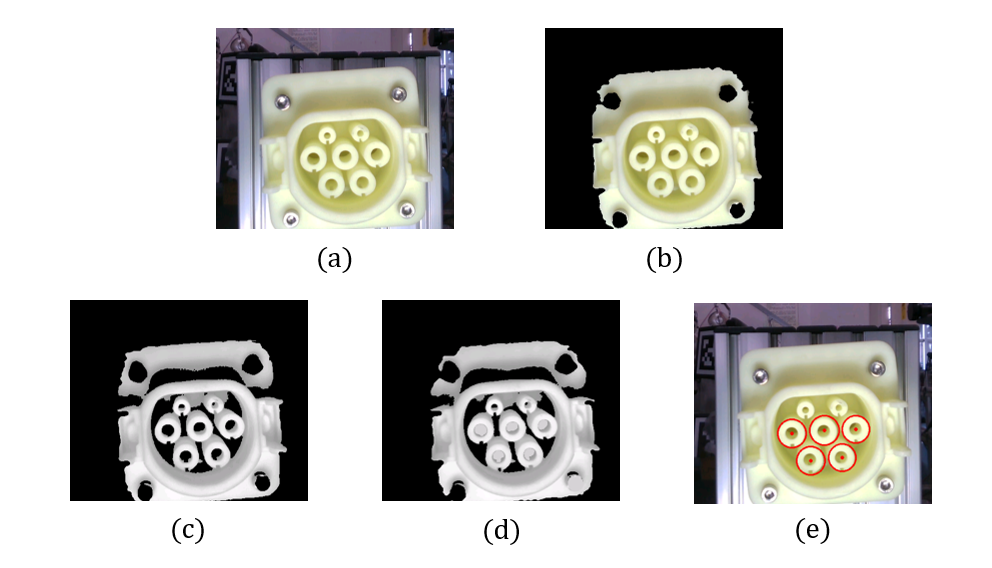}
\caption{Feature points extraction pipeline. Fig.(a) is the raw image from the camera. Fig.(b) is the extraction of charging port based on HSV information. Fig.(c) shows the erosion operation after converting the extracted charging port into a gray image. Fig.(d) is the first stage of the circle identification. We fill the extracted circle centers for more accurate circle extraction. Fig.(e) is the final extraction of features using the Hough transform. }
\label{visual_servo}
\end{figure}

\begin{figure}[t]
\centering
\subfigure[]{
\includegraphics[width=7.1cm]{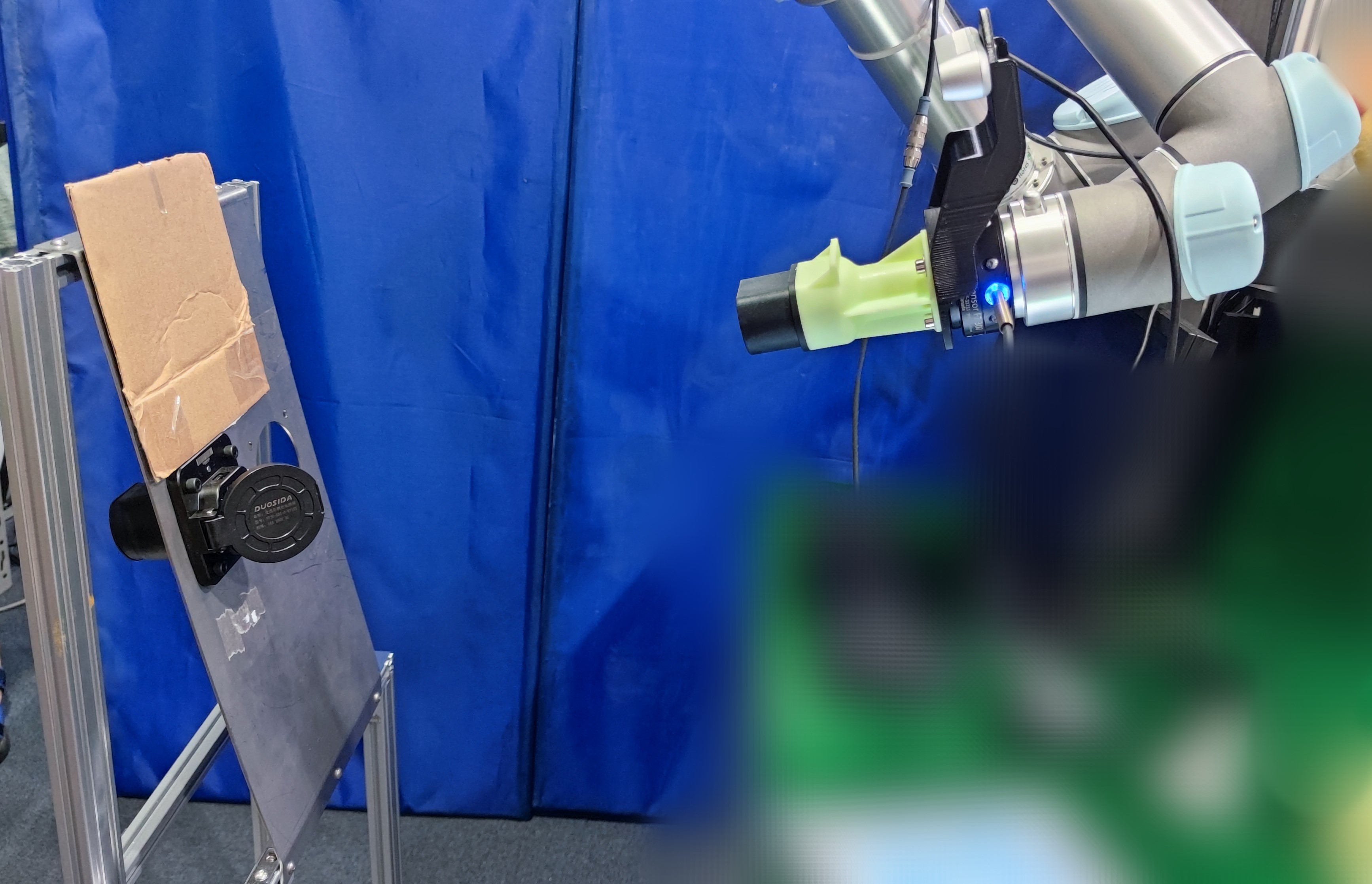}
}%
\quad
\subfigure[]{
\includegraphics[width=3.5cm]{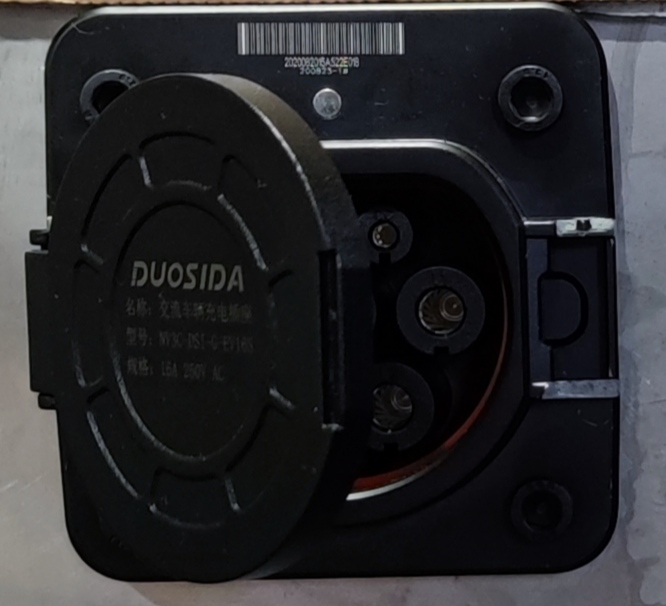}
}%
\subfigure[]{
\includegraphics[width=3.5cm]{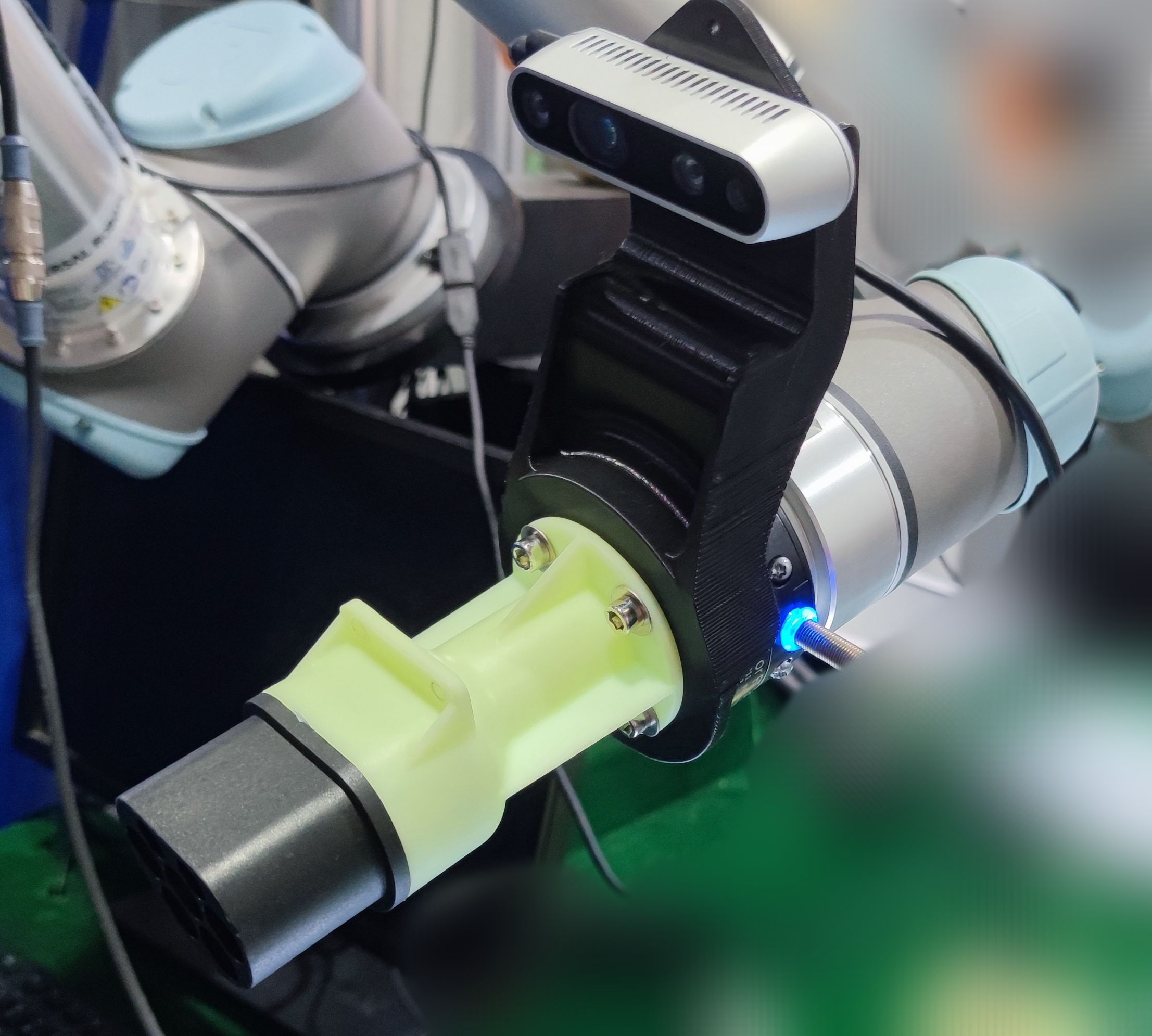}
}%
\caption{The real experiment configuration. Fig.(a) is the overview of the scenario. Fig.(b) is the charging port. Fig.(c) is the sensors and charger attached to the end of UR5.}
\label{real_scene}
\end{figure}
\section{Experiments}
We implement and verify the feasibility of our method in both the simulation and the real world. We use UR5 robot arm with the robotiq ft300 F/T sensor and the Intel D435i RGB-D camera in our experiment. The camera and the F/T sensor are fixed at the robot wrist by 3d-printed adapter as shown Fig.~\ref{real_scene}.

\subsection{Charging Cover Perception and Manipulation}
At first we obtain a point cloud of the world, and then roughly crop the point cloud to keep the plane and charging port in it. Before the segmentation, we set $K=2$ in K-means, that means to divide the point cloud into two parts, one for plane and charging port and the other for the cover. The area of the cover is significantly small than the plane, so we choose the point cloud contained less points to represent the cover. Then we filter the point cloud to eliminate isolated points. Finally we get the center and the normal of the cover. These processes are illustrated in Fig.~\ref{esti}.

Next we test the manipulation method. We measured the $r_c=4cm$, and we set the threshold of force $\left \| F \right \| \le 3N$ as the condition for charger touching the charger in the $attempt$ stage and stopping the charger in the $open$ stage.

We set up five sets of experiments with the cover angles in $[15^{\circ},30^{\circ},45^{\circ},60^{\circ}, 75^{\circ}]$, and the angle is obtained with the help of AprilTag. We've found that in the experiment with the degree $15^{\circ}$, we can not distinguish between the plane and the cover, so we failed in this situation. In the other four sets of experiments, we reached the $x_e=[-2.1, 1.3, 3.4, -0.8](cm)$ in $attempt$ stage, and in all of them we successfully opened the cover. This proves the necessity of the $attempt$ stage and the effectiveness of our method.

\begin{figure}[t]
\centerline{\includegraphics[width=2.2in]{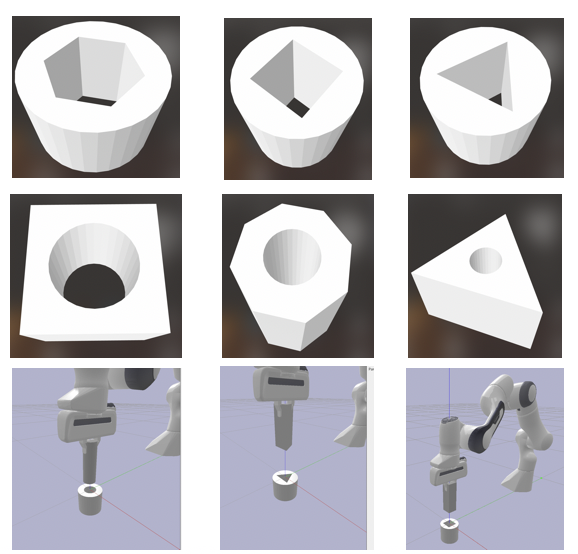}}
\caption{Different hole geometries for training}
\label{fiiig}
\end{figure}
\subsection{Peg-in-Hole}
In our experiment, the charging port we use has five obvious circles on the surface. Different features can be selected accordingly in other scenarios. Then we extracted these circles by the HSV and take the centers of these circles as the feature points. The feature extraction pipeline is shown in Fig.~\ref{visual_servo}. According the feature points, we get the velocity $v_{we}$ of the charger as described from (\ref{visio1}) to (\ref{visio5}), and then the charger is controlled to approach the charging port by the visual servo.

In the insert stage, we firstly train the hybrid position-force controller in the PyBullet \cite{coumans2021} simulation and then transfer the model directly to the real world without any fine-tuning. In order to eliminate the simulation-to-reality gap, we train the model with different hole geometries parallel to possibly cover the complete force distribution in different hole shapes as shown in Fig.~\ref{fiiig}.

\begin{figure}[tbp]
\centerline{\includegraphics[width=3in]{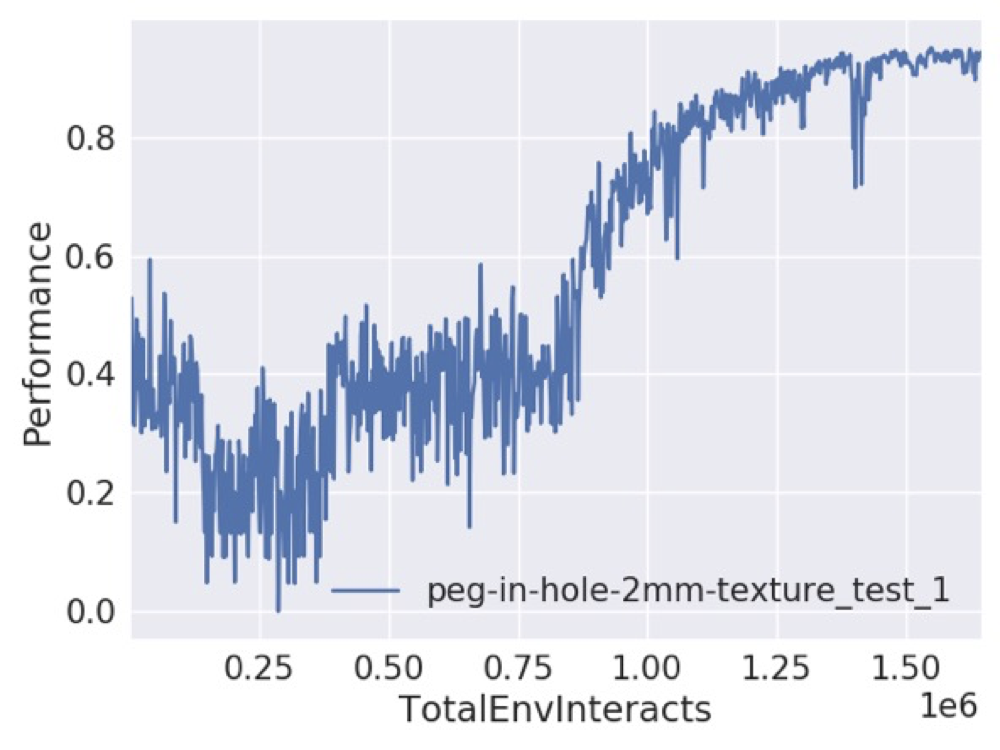}}
\caption{Average reward of reinforcement learning training}
\label{curve}
\end{figure}

We use A2C, a modern RL algorithm for the hybrid position-force controller. The model is trained in PyBullet for about 5 hours. The average reward starts to converge in 1 million interaction with the environment as shown in Fig.~\ref{curve}.

We compare the proposed method for peg-in-hole task with the traditional methods, which are random search and spiral search. Random search and spiral search are based on pure force control. For this reason, the initial position error is limited to 5mm and we assume there is no orientation error. As shown in \ref{table}, there are notable improvements in both accuracy and efficiency compared with the two traditional methods even when our initial error is larger than the traditional ones.

\begin{table}[t]
\caption{Peg-in-Hole Benchmark} 
\centering 
\begin{tabular}{|c| c c c|} 
\hline\hline 
  & Random search & Spiral search & Proposed \\ [0.5ex] 
\hline 
position error & 5mm & 5mm & 500mm \\
orientation error & -- & -- & 2$\pi$ \\
accuracy & 0.04 & 0.27 & 0.97 \\ 
time & 15s & 10s & 9s \\
\hline 
\end{tabular}
\label{table} 
\end{table}

\section{Conclusions}
We present a complete method for the automatic vehicle charging system. With vision-force fusion we obtain the rapid and accurate perception results. The whole system is transformed from simulation to the real world without any fine-tuning, and is testified in the experiments we design. In our system, the robot arm is fixed, which limits the location of the charging port. In the future research, the system will contain an mobile robot arm so that it can finish the charging task wherever the electric vehicle is.




%
%

\bibliographystyle{ieeetr}
\bibliography{ref.bib}

\end{document}